\documentclass[letterpaper]{article} 
\usepackage[]{aaai23}  
\usepackage{times}  
\usepackage{helvet}  
\usepackage{courier}  
\usepackage[hyphens]{url}  
\usepackage{graphicx} 
\usepackage{natbib}  
\usepackage{caption} 
\usepackage{algorithm}
\usepackage{algorithmic}

\graphicspath{ {./figures/} }

\usepackage{newfloat}
\usepackage{listings}
\floatstyle{ruled}
\newfloat{listing}{tb}{lst}{}
\floatname{listing}{Listing}

\usepackage{tikz}
\tikzset{>=latex}
\usepackage[utf8]{inputenc}
\usepackage{amsmath,amsfonts,amsthm,amssymb,physics}
\usepackage{graphicx}
\usepackage{url}
\usepackage{color}
\usepackage[colorlinks=true, linkcolor=red, urlcolor=blue, citecolor=blue]{hyperref}
\usepackage{booktabs}
\usepackage{algorithm}
\usepackage{algorithmic}
\usepackage{bm}
\usepackage{stackrel}
\usepackage{enumitem}
\usepackage{breqn}
\usepackage{booktabs}
\usepackage{siunitx}

\usepackage[parfill]{parskip}

\newcommand{\Ebb}{{\mathbb{E}}}

\newcommand{\Nbb}{{\mathbb{N}}}

\newcommand{\Rbb}{{\mathbb{R}}}

\usepackage{etoolbox}
\renewcommand{\bfseries}{\fontseries{b}\selectfont} 
\robustify\bfseries             
\newrobustcmd{\B}{\bfseries}

\title{Wildfire Forecasting with Satellite Images and Deep Generative Model}
\author {
    Thai Nam Hoang\textsuperscript{\rm 1, 2},
    Sang T. Truong\textsuperscript{\rm 3},
    Chris Schmidt\textsuperscript{\rm 2}
}
\affiliations {
    \textsuperscript{\rm 1} Beloit College \\
    \textsuperscript{\rm 2} SSEC, University of Wisconsin, Madison \\
    \textsuperscript{\rm 3} Stanford University \\
    thoang5@wisc.edu, sttruong@cs.stanford.edu, chris.schmidt@ssec.wisc.edu
}

\usepackage{bibentry}

\begin{document}

\maketitle

\begin{abstract}
Wildfire forecasting has been one of the most critical tasks that humanities want to thrive. It plays a vital role in protecting human life. Wildfire prediction, on the other hand, is difficult because of its stochastic and chaotic properties. We tackled the problem by interpreting a series of wildfire images as a video and used it to anticipate how the fire would behave in the future. However, creating video prediction models that account for the inherent uncertainty of the future is challenging. The bulk of published attempts is based on stochastic image-autoregressive recurrent networks, which raises various performance and application difficulties, such as computational cost and limited efficiency on massive datasets. Another possibility is to use entirely latent temporal models that combine frame synthesis and temporal dynamics. However, due to design and training issues, no such model for stochastic video prediction has yet been proposed in the literature. This paper addresses these issues by introducing a novel stochastic temporal model whose dynamics are driven in a latent space. It naturally predicts video dynamics by allowing our lighter, more interpretable latent model to beat previous state-of-the-art approaches on the GOES-16 dataset. Results will be compared towards various benchmarking models.

\end{abstract}

\section{Introduction}
Weather forecasting has been one of the essential tasks of humankind. Since 1975, the US Government has launched the Geostationary Operational Environmental Satellite (GOES) to produce data that can enhance weather and climate models, thus allowing for more precise and quick weather forecasting and a better knowledge of long-term climate \cite{goes_envi_sat, goes_history}. Using informative GOES data can help with wildfire prediction and detection, as lately, more and more wildfires are happening. The increase in severity, frequency, and duration of wildfires brought on by anthropogenic climate change and rising global temperatures has resulted in the emission of significant amounts of greenhouse gases, the destruction of forests and their related habitats, and damage to infrastructure and property \cite{marlon2012, abatzoglou2016, vila2020}. 

Previous projects on fire detection have been done. However, most of them tried to tackle segmentation \cite{khryashchev2020, pereira2021, zhang2021}. However, these methods may not serve the whole idea of wildfire prediction and prevention, as they tend to recognize the fire once it happens already, and the larger it grows, the easier they can segment. Additionally, they only tried to use traditional rasterized satellite images, where images are patched into traditional RGB 3-channel. Therefore they could not capture very early factors of small wildfire. 

An advantage of GOES compared to other satellites is that GOES uses an Advanced Baseline Imager (ABI), which takes the image of the Earth with 16 spectral bands (two visible channels, four near-infrared channels, and ten infrared channels) with a fast scan time of 12 slices per hour and higher resolution of 0.5-2km \cite{schmit2017, abi}. We can utilize the robustness of ABI images to create a temporal-like dataset that serves as baseline data for prediction. 

For the sake of synthesizing images, generating adversarial networks (GANs) can be taken into account. The network introduces a generator and discriminator for unsupervised adversarial training, which indirectly “learns” the dataset through a minimax game \cite{goodfellow2014}. The discriminator distinguishes between genuine pictures from a training set and synthetic phony ones created by the generator. Starting from this idea, we can evolve to generate video frames instead of a single image. This idea can be classified as stochastic video prediction. However, it is a daunting task as most approaches are usually based on image-autoregressive models \cite{babaeizadeh2017, denton2018, weissenborn2019}, which was a pixel-wise tackle and built around Recurrent Neural Networks (RNNs), where each generated frame is fed back to the model to produce the next frame. However, the performance of this approach relies heavily on the capability of its encoders and decoders, as each generated frame has to be re-encoded in a latent space. Such techniques may have a negative impact on performance and have limited application, especially when dealing with massive amounts of data \cite{gregor2018, rubanova2019}.

Another technique is to separate the dynamic of the state representations from the produced frames, which are decoded separately from the latent space. This is computationally interesting when combined with a low-dimensional latent space and eliminates the relationship mentioned above between frame generation and temporal dynamics. Furthermore, such models are more interpretable than autoregressive models and may be used to create a complete representation of a system's state, for example, in reinforcement learning applications \cite{gregor2018}. However, these State-Space Models (SSMs) are more challenging to train since they need non-trivial inference systems \cite{krishnan2016} and careful dynamic model construction \cite{karl2016}. As a result, most effective SSMs are only assessed on minor or contrived toy tasks.

In this paper, we present a novel stochastic dynamic model for video prediction that successfully harnesses the structural and computational advantages of SSMs operating on low-dimensional latent spaces. Its dynamic component governs the system's temporal evolution through residual updates of the latent state, which are conditioned on learned stochastic variables. This approach enables us to execute an efficient training strategy and analyze complex high-dimensional data such as movies in an interpretable way. This residual principle is related to recent breakthroughs in the relationship between residual networks and Ordinary Differential Equations (ODEs). As illustrated in our research, this interpretation offers additional possibilities, such as creating videos at varied frame rates. As evidenced by comparisons with competing baselines on relevant benchmarks, the proposed technique outperforms existing state-of-the-art models on the task of stochastic video prediction.

\section{Related Works}
Video synthesis encompasses a wide range of tasks, from super-resolution \cite{caballero2016}, interpolation between
distant frames \cite{jiang2017}, generation \cite{tulyakov2017}, video-to-video translation \cite{wang2018}, and conditioning video prediction, which is the subject of this study.

\subsection{Deterministic models}
Beginning with RNN-based sequence generating models \cite{graves2013}, a variety of video prediction algorithms based on LSTMs (Long Short-Term Memory networks \cite{lstm}) and its convoluted variation of ConvLSTMs \cite{shi2015} were developed \cite{srivastava2015, brabadere2016, wichers2018, jin2020}.
Indeed, computer vision algorithms are frequently aimed at high-dimensional video sequences and employ domain-specific approaches such as pixel-level transformations, and optical flow \cite{walker2015, walker2016, vondrick2017, luchaochao2017, fan2019} to help in the generation of high-quality predicting outputs. However, such algorithms are deterministic, which limits their efficacy by failing to produce high-quality long-term video frames \cite{babaeizadeh2017, denton2018}. Another approach is to apply adversarial losses \cite{goodfellow2014} to sharpen the resulting frames \cite{vondrick2017, luchaochao2017, wu2020}. Adversarial losses, on the other hand, are famously challenging to train, and as a result, mode collapse develops, restricting generational diversity.

\subsection{Stochastic and image-autoregressive models}
Other methods manipulate pixel-level autoregressive generation and concentrate on precise probability maximization \cite{oord2016, kalchbrenner2016, weissenborn2020}. Flow normalization has also been studied using invertible transformations between the observation and latent spaces \cite{kingma2018, kumar2019}. However, they necessitate the careful construction of sophisticated temporal production systems that manage high-dimensional data, resulting in exorbitant temporal generation costs. For the inference of low-dimensional latent state variables, Variational Auto-encoders are utilized in more efficient continuous models (VAEs \cite{kingma2013}).Stochastic variables were integrated into ConvLSTM in \cite{babaeizadeh2017}. In order to sample random variables that are supplied to a predictor LSTM, both \cite{he2018} and \cite{denton2018} utilized a prior LSTM conditioned on previously produced frames. Finally, \cite{lee2018} merged the ConvLSTM with learned prior, sharpening the resulting videos with an adversarial loss. However, all of these approaches are image-autoregressive in that they feed their predictions back into the latent space, connecting the frame synthesis and temporal models together and increasing their computing cost. \cite{minderer2019} offered an autoregressive VRNN model based on learned image key points rather than raw frames, which is similar to our work. It is unknown to what degree this adjustment will alleviate the issues mentioned above. Instead, we address these concerns by focusing on video dynamics and proposing a state-space model that operates on a limited latent space.

\subsection{State-space model}
Numerous latent state-space models, often trained using deep variational inference, have been suggested for sequence modelization \cite{bayer2014, fraccaro2016, hafner2018}. These initiatives, which employ locally linear or RNN-based dynamics, are intended for low-dimensional data since learning such models on complex data is difficult or concentrates on control or planning tasks. On the other hand, the utterly latent technique is the first to successfully apply to complex high-dimensional data such as videos due to a temporal model based on residual updates of its latent state. It is part of a recent development that connects differential equations with neural networks \cite{lu2017, long2017}, leading to the integration of ODEs, which are seen as continuous residual networks \cite{resnet}. On the other hand, follow-ups and similar research are confined to low-dimensional data, prone to overfitting, and unable to manage stochasticity within a sequence. Another line of research examines stochastic differential equations using neural networks \cite{ryder2018, debrouwer2019}, but is confined to continuous Brownian noise, whereas video generation also involves modeling of punctual stochastic events.

\section{Methods}
We are concerned with the challenge of stochastic video prediction, attempting to forecast future frames from a video given the initial conditioning frames.

\subsection{Latent Residual Dynamic Model}
Let $x_{0:T} = \{x_{0}, x_{1}, \ldots, x_{T - 1}\}$ be a sequence of $T$ video frames, where each state $x_{t} \in \Rbb^{b \times m \times n}$ is a satellite image, where $b = 10$ is the number of band ("channel"), and $m = 1500$ and $n = 2500$ is the maximum image size. We want generate $x_{T: T + h}$, where $h$ is the forecasting horizon. One way to achieve this goal is to use an parameterized autoregressive model $f_{\theta}$ that maps one state to another: $x_{t + 1} = f_{\theta}(x_{t})$. We introduce latent variables $y$ that a dynamic temporal model drives to achieve this. Each frame $x_{t}$ is then generated from the corresponding latent state $y_{t}$ only, making the dynamics independent from the previously generated frames.

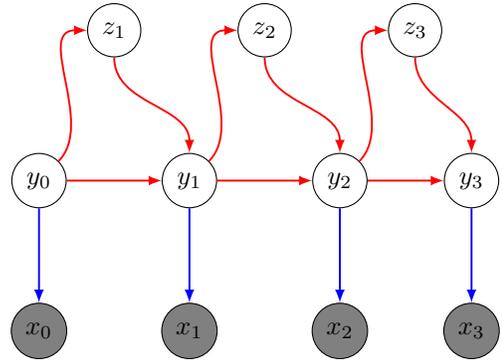
\begin{figure}[t]
    \centering
    \begin{tikzpicture}[
            unobserved/.style={fill=white, draw=black, shape=circle},
            observed/.style={fill=gray, draw=black, shape=circle},
            edge_red/.style={->, fill=none, draw=red, line width=0.25mm},
            edge_blue/.style={->, fill=none, draw=blue, , line width=0.25mm},
        ]
    \pgfdeclarelayer{nodelayer}
    \pgfdeclarelayer{edgelayer}
    \pgfsetlayers{nodelayer, edgelayer}
    	\begin{pgfonlayer}{nodelayer}
    		\node [style=observed] (0) at (-5, 1) {$x_{0}$};
    		\node [style=observed] (1) at (-3, 1) {$x_{1}$};
    		\node [style=observed] (2) at (-1, 1) {$x_{2}$};
    		\node [style=observed] (3) at (0.75, 1) {$x_{3}$};
    		\node [style=unobserved] (5) at (-5, 3) {$y_{0}$};
    		\node [style=unobserved] (14) at (-4, 5) {$z_{1}$};
    		\node [style=unobserved] (15) at (-3, 3) {$y_{1}$};
    		\node [style=unobserved] (26) at (-2, 5) {$z_{2}$};
    		\node [style=unobserved] (27) at (-1, 3) {$y_{2}$};
    		\node [style=unobserved] (30) at (0, 5) {$z_{3}$};
    		\node [style=unobserved] (31) at (0.75, 3) {$y_{3}$};
    	\end{pgfonlayer}
    	\begin{pgfonlayer}{edgelayer}
    		\draw [style={edge_blue}] (5) to (0);
    		\draw [style={edge_blue}] (15) to (1);
    		\draw [style={edge_blue}] (27) to (2);
    		\draw [style={edge_blue}] (31) to (3);
    		\draw [style={edge_red}, in=-180, out=45] (5) to (14);
    		\draw [style={edge_red}, in=90, out=-90] (14) to (15);
    		\draw [style={edge_red}, in=180, out=45] (15) to (26);
    		\draw [style={edge_red}, in=90, out=-90] (26) to (27);
    		\draw [style={edge_red}, in=-180, out=45] (27) to (30);
    		\draw [style={edge_red}, in=90, out=-90] (30) to (31);
    		\draw [style={edge_red}] (5) to (15);
    		\draw [style={edge_red}] (15) to (27);
    		\draw [style={edge_red}] (27) to (31);
    	\end{pgfonlayer}
    \end{tikzpicture}
    \caption{Generative model $p$}
    \label{fig:1-generative}
\end{figure}

Based on \cite{srvp}, we suggest using a stochastic residual network to describe the transition function of the latent dynamic of $y$. State $y_{t+1}$ is selected to be deterministically dependent on the preceding state $y_{t}$ and conditionally dependent on an auxiliary random variable $z_{t+1}$. These auxiliary variables represent the video dynamics' unpredictability. They have a learned factorized Gaussian prior that is solely affected by the initial state. The model is depicted in Figure (\ref{fig:1-generative}) and defined as follows:

\begin{equation}
\label{eq:1}
\begin{cases}
    y_{1} \sim \mathcal{N}(0, I), \\
    z_{t + 1} \sim \mathcal{N}(\mu_{\theta}(y_{t}), \sigma_{\theta}(y_{t})I), \\
    y_{t + 1} = y_{t} + f_{\theta}(y_{t}, z_{t + 1}), \\
    x_{t} \sim \mathcal{G}(g_{\theta}(y_{t}))
\end{cases}
\end{equation}

where $\mu_{\theta}, \sigma_{\theta}, f_{\theta}, g_{\theta}$ are neural nets, and $\mathcal{G}(g_{\theta}(y_{t}))$ is probability distribution parameterized by $g_{\theta}(y_{t})$. Note that $y_{1}$ is assumed to have a standard Gaussian prior and, in our VAE setting, will be inferred from conditioning frames for the prediction.

\begin{figure}[t]
    \centering
    \begin{tikzpicture}[
            unobserved/.style={fill=white, draw=black, shape=circle},
            observed/.style={fill=gray, draw=black, shape=circle},
            edge_red/.style={->, fill=none, draw=red, line width=0.25mm},
            edge_blue/.style={->, fill=none, draw=blue, , line width=0.25mm},
            path_red/.style={-, draw=red, line width=0.25mm}
        ]
    \pgfdeclarelayer{nodelayer}
    \pgfdeclarelayer{edgelayer}
    \pgfsetlayers{nodelayer, edgelayer}
        \begin{pgfonlayer}{nodelayer}
    		\node [style=observed] (3) at (-3, -2) {$x_{0}$};
    		\node [style=unobserved] (6) at (-2, 2) {$z_{1}$};
    		\node (11) at (-2, 0.25) {};
    		\node [style=unobserved] (27) at (-3, 0) {$y_{0}$};
    		\node [style=unobserved] (28) at (-1, 0) {$y_{1}$};
    		\node [style=unobserved] (29) at (1, 0) {$y_{2}$};
    		\node [style=unobserved] (30) at (0, 2) {$z_{2}$};
    		\node [style=unobserved] (33) at (3, 0) {$y_{3}$};
    		\node [style=unobserved] (43) at (2, 2) {$z_{3}$};
    		\node [style=observed] (45) at (3, -2) {$x_{3}$};
    		\node [style=observed] (46) at (-1, -2) {$x_{1}$};
    		\node (47) at (0, 0.25) {};
    		\node [style=observed] (51) at (1, -2) {$x_{2}$};
    		\node (52) at (2, 0.25) {};
	    \end{pgfonlayer}
	    \begin{pgfonlayer}{edgelayer}
    		\draw [style={edge_red}, in=240, out=105, looseness=1.25] (11.center) to (6);
    		\draw [in=-75, out=60, looseness=1.25] (3) to (11.center);
    		\draw [style={edge_blue}] (3) to (27);
    		\draw [style={path_red}, in=-75, out=60, looseness=1.25] (3) to (11.center);
    		\draw [in=-75, out=60, looseness=1.25] (46) to (47.center);
    		\draw [style={path_red}, in=-75, out=60, looseness=1.25] (46) to (47.center);
    		\draw [style={path_red}, in=-75, out=60, looseness=1.25] (51) to (52.center);
    		\draw [style={edge_red}, in=240, out=105, looseness=1.25] (47.center) to (30);
    		\draw [in=-90, out=30] (3) to (47.center);
    		\draw [style={path_red}, in=-90, out=30] (3) to (47.center);
    		\draw [style={edge_red}, in=240, out=105, looseness=1.25] (52.center) to (43);
    		\draw [in=-90, out=30] (46) to (52.center);
    		\draw [style={path_red}, in=-90, out=30] (46) to (52.center);
    		\draw [style={path_red}, in=-75, out=120, looseness=1.25] (45) to (52.center);
    		\draw [style={edge_blue}, in=-45, out=180, looseness=1.25] (46) to (27);
    		\draw [style={path_red}, in=-75, out=120] (46) to (11.center);
    		\draw [style={path_red}, in=-75, out=60, looseness=1.25] (46) to (47.center);
    		\draw [style={path_red}, in=-75, out=120, looseness=1.25] (51) to (47.center);
    		\draw [style={path_red}, in=-105, out=30, looseness=0.75] (3) to (52.center);
	    \end{pgfonlayer}
    \end{tikzpicture}
    \caption{Inference model $q$}
    \label{fig:2-inference}
\end{figure}
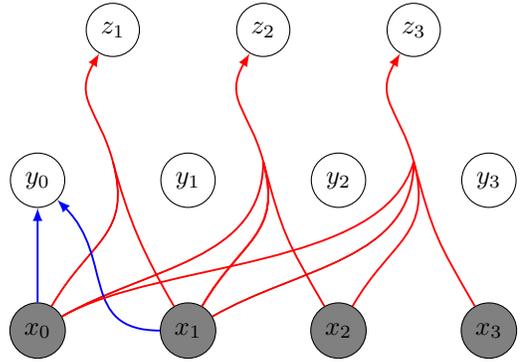

The residual update rule is based on the Euler differentiation technique for differential equations. The state of the system $y_{t}$ is updated by its first-order movement, i.e, the residual $f_{\theta}(y_{t}, z_{t + 1})$. +1). This fundamental idea makes our temporal model lighter and more interpretable than a normal RNN. Equation (\ref{eq:1}), on the other hand, differs from a discretized ODE because of the introduction of the stochastic discrete-time variable $z$. Nonetheless, we recommend that the Euler step size $\Delta t$ always be less than $1$ to get the temporal model closer to continuous dynamics.
With $\frac{1}{\Delta t} \in \Nbb$ to synchronize the step size with the video frame rate, the updated dynamics are as follows:

\begin{equation}
\label{eq:2}
y_{t + \Delta t} = y_{t} + \Delta {t} \cdot f_{\theta}(y_{t}, z_{\lfloor t\rfloor + 1})
\end{equation}

The auxiliary variable $z_{t}$ is held constant between two integer time steps in the formulation. It should be noted that during training or testing, a different $\Delta t$ might be utilized. Because each intermediate latent state may be decoded in the observation space, our model can generate videos at any frame rate. This capacity allows us to assess the learned dynamic's quality while challenging its ODE inspiration by evaluating its generalization to the continuous limit. For the rest of this section, we consider that $\Delta t = 1$; one can generalize to a smaller $\Delta t$ is straightforward as Figure (\ref{fig:1-generative}) remains unchanged.

\subsection{Content Variable}\label{sec:content-var}
Some video sequence components, such as the terrain, can be static or vary very slightly, such as the moving clouds. They may not affect the dynamics; thus, we model them separately, as \cite{denton2017, yingzhen18a} have done. We calculate a content variable $w$ that remains constant throughout the generation process and is passed into the frame generator along with $y_{t}$.
It allows the dynamical element of the model to concentrate just on movement, making it lighter and more stable. Furthermore, it enables us to use architectural developments in neural networks, such as skip connections \cite{ronneberger2015}, to generate more realistic frames.

This content variable is a deterministic function $c_{\psi}$ of a fixed number $k < T$ of frames $x_{c}^{(k)} = (x_{i_{0}}, x_{i_{2}}, \ldots, x_{i_{k}})$:
\begin{equation}
\label{eq:3}
\begin{cases}
    w = c_{\psi}(x_{c}^{(k)}) = c_{\psi}(x_{i_{0}}, x_{i_{2}}, \ldots, x_{i_{k}}) \\
    x_{t} \sim \mathcal{G}(g_{\theta}(y_{t}))
\end{cases}
\end{equation}

During testing, $x_{c}^{(k)}$ are the last $k$ conditioning frames. 

In contrast to the dynamic variables $y$ and $z$, this content variable has no probabilistic prior. As a result, the information it carries is only structurally confined rather than in the loss function. To avoid leaking temporal information in $w$, we propose sampling these k frames evenly inside $x_{0:T}$ during training. We also build $c_{\psi}$ as a permutation invariant function \cite{zaheer2017} composed of an MLP fed with the sum of each frame representation, as shown in \cite{santoro2017}.

Because of the absence of prior and architectural limitations, $w$ can contain as much non-temporal information as feasible while excluding dynamic information. On the other hand, $y$ and $z$ should only include temporal information that $w$ cannot capture owing to their high standard Gaussian priors.

This content variable may be deleted from our model, resulting in a more traditional deep state-space model.

\subsection{Variational Inference and Architecture}
Following the generative model depicted in Figure (\ref{fig:1-generative}), the conditional join probability of the full model, given a content variable $w$, can be written as:

\begin{equation}
\label{eq:4}
\begin{split}
    & p(x_{0:T}, z_{1:T}, y_{0:T} | w) \\
    = \;\; & p(y_{0})\prod_{t = 1}^{T}p(z_{t}, y_{t} | y_{t - 1})\prod_{t = 0}^{T}p(x_{t}|y_{t}, w)
\end{split}
\end{equation}

with

\begin{equation}
\label{eq:5}
p(z_{t}, y_{t} | y_{t - 1}) = p(z_{t}|y_{t - 1})p(y_{t}|y_{t - 1}, z_{t})
\end{equation}

According to Equation (\ref{eq:1}), $p(y_{t} | y_{t - 1}, z_{t}) = \delta(y_{t} - y_{t - 1} - f_{\theta}(y_{t - 1}, z_{t}))$, where $\delta$ is the Dirac delta function centered at $0$. Hence, in order to optimize the likelihood of the observed videos $p(x_{0:T}|w)$, we need to infer latent variables $y_{0}$ and $z_{1:T}$. This can be done by deep variational inference using the inference model parameterized by $\phi$ and shown in Figure (\ref{fig:2-inference}), which comes down to considering a variational distribution $q_{Z, Y}$ defined and factorized as follows:

\begin{equation}
\label{eq:6}
\begin{split}
    & q_{Z, Y} \triangleq q(z_{1:T}, y_{0:T}|x_{0:T}, w) \\
    = \;\; & q(y_{0}|x_{0:k})\prod_{t = 1}^{T}q(z_{t}|x_{0:t})q(y_{t}|y_{t - 1}, z_{t}) \\
    = \;\; & q(y_{0}|x_{0:k})\prod_{t = 1}^{T}q(z_{t}|x_{0:t})p(y_{t}|y_{t - 1}, z_{t})
\end{split}
\end{equation}

where $q(y_{t}|y_{t - 1}, z_{t}) = p(y_{t}|y_{t - 1}, z_{t})$ begin the aforementioned Dirac delta function. This yields the following evidence lower bound (ELBO):

\begin{equation}
\label{eq:7}
\begin{split}
    & \log p(x_{0:T}|w) \geq \mathcal{L}(x_{0:T}; w, \theta, \phi) \\
    \triangleq \;\; & -D_{\text{KL}}[q(y_{0}|x_{0:k}) \parallel p(y_{0})] \\
    & + \Ebb_{(\tilde{z}_{1:T}, \tilde{y}_{0:T}) \sim q_{Z, Y}} \bigg[\sum_{t = 0}^{T} \log p(x_{t}|\tilde{y}_{t}, w) \\
    & - \sum_{t = 1}^{T} D_{\text{KL}}[q(z_{t}|x_{0:t}) \parallel p(z_{t}|\tilde{y}_{t - 1}]\bigg]
\end{split}
\end{equation}

where $D_{\text{KL}}$ denotes the Kullback-Leibler (KL) divergence.

The sum of KL divergence expectations implies considering the full past sequence of inferred states for each time step due to the dependence on conditionally deterministic variable $y_{1:T}$. However, optimizing $\mathcal{L}(x_{0:T}; w, \theta, \psi)$ with respect to model parameter $\theta$ and variational parameters $\phi$ can be done efficiently by sampling a single full sequence of states from $q_{Z, Y}$ per example, and computing gradients by backpropagation \cite{rumelhart1986}, trough all inferred variables, using reparameterization trick \cite{kingma2013}.
We classically choose $q(y_{0}|x_{0:k})$ and $q(z_{t}|x_{0:t})$ to be factorized Gaussian so that all KL divergences can be computed analytically.

We include an $L_{2}$ regularization term on residual $f_\theta$ applied to $y$, which stabilizes the temporal dynamics of the residual network, as noted by \cite{behrmann2018, bezenac2019, rousseau2019}. Given a set of videos $\mathcal{X}$, the complete optimization problem, where $\mathcal{L}$ is defined as in Equation (\ref{eq:7}), is then given as:

\begin{equation}
\label{eq:8}
\begin{split}
    \arg\max_{\theta, \phi, \psi} & \sum_{x \in \mathcal{X}}\bigg[ \Ebb_{x_{c}^{(k)}}\mathcal{L}(x_{0:T};c_{\psi}(x_{c}^{(k)}), \theta, \phi) \\
    & -\lambda\Ebb_{(\tilde{z}_{1:T}, \tilde{y}_{0:T}) \sim q_{Z, Y}} \sum_{t = 1}^{T} \parallel f_{\theta} (y_{t - 1, z_{t}}) \parallel_{2}\bigg]
\end{split}
\end{equation}

The first latent variables are inferred with the conditioning framed and are then predicted with the dynamic model. In contrast, each frame of the input sequence is considered for inference during training, which is done as follows. Firstly, each frame $x_{t}$ is independently encoded into a vector-valued representation $\tilde{x}_{t}$, with $\tilde{x}_{t} = h_{\psi}(x_{t})$. $y_{0}$ is then inferred using an MLP on the first $k$ encoded frames $\tilde{x}_{0:k}$. Each $z_{t}$ is inferred in a feed-forward fashion with an LSTM on the encoded frames. Inferring $z$ this way experimentally performs better than, e.g., inferring them from the whole sequence $x_{0:T}$; we hypothesize that this follows from the fact that this filtering scheme is closer to the prediction setting, where the future is not available.

\section{Experiments}

\subsection{Training}
In this part, we qualitatively investigate the dynamics and latent space learned by our model. 

The stochastic nature and originality of the video prediction task make it challenging to evaluate ordinarily \cite{lee2018}: because the task is stochastic, comparing the ground truth and a predicted video is insufficient. We, therefore, follow the standard strategy \cite{denton2018, lee2018}, which consists of sampling a specific number (here, 100 samples) of probable futures from the tested model and reporting the highest performing sample against the genuine video for each test sequence. We show this disparity for two generally used metrics that are computed frame-by-frame and averaged over time: Peak Signal-to-Noise Ratio (PSNR, \emph{higher is better}) and Structured Similarity (SSIM, \emph{higher is better}) \cite{hore2010}. PSNR penalizes inaccuracies in projected dynamics since it is a pixel-wise measurement derived from $L 2$ distance, but it may also favor fuzzy predictions. To avoid this problem, SSIM compares local frame patches, although this results in some dynamics information being lost. We consider these two measures complementary since they capture distinct sizes and modalities.

We present experimental results on the GOES data that we briefly present in the following section. We also compare our model against SVG \cite{denton2018} and StructVRNN \cite{minderer2019}. SVG has the most similar training and architecture among the models. To do fair comparisons using this technique, we employ the same neural architecture as SVG for our encoders and decoders. Unless otherwise indicated, our model is evaluated with the same $\Delta t$ as in training, as shown in Equation (\ref{eq:2}).

\subsection{Dataset}
We used GOES-16 CONUS (ABI-L1b-RadC) infrared ABI spectral bands (band 7 - 16) \cite{abi_guide}, specifically at night from 22:00:00 to 5:00:00 CST (UTC-5) or 5:00:00 to 12:00:00 UTC. The dataset can be pulled directly from AWS s3 \cite{noaa_goes16}. We set the data from day 80 to day 135 of the year 2022, which is 56 days or eight weeks. For each band, there will be a total of 4704 slices.

\subsection{Preprocessing}
Initially, each slice will be $1500 \times 2500$. Since a single slice sizes about 3-4Mb, we decided to crop to $256 \times 256$ to, first of all, cut down the size and, therefore, speed up the calculation. Moreover, doing so will center our focus on the Mid and South regions of the US, where wildfires usually happen during spring and summer. Each pixel represents the value of radiance of brightness in each band. We first have to convert the other band's radiance temperature into band 7's radiance temperature. We first calculate its brightness temperature by applying the Planck function, and the spectral bandpass correction into radiance temperature \cite{noaa_formula}:

\begin{equation}
    BT = \bigg[\frac{fk_{2}}{\log(\frac{fk_{1}}{L_{v}} + 1)} - bc_{1}\bigg] \times \frac{1}{bc_{2}}
\end{equation}

where $L_v$ is the radiance, $fk_{1}$ and $fk_{2}$ are coefficients of the Planck function derived from physical constants (i.e., the speed of light, the Boltzmann constant, and the Planck constant) and the bandpass central wavenumber, and $bc_{1}$ and $bc_{2}$ are the spectral response function offset and scale correction terms. These four coefficients are included in the product metadata as variables: \lstinline{planck_fk1}, \lstinline{planck_fk2}, \lstinline{planck_bc1}, and \lstinline{planck_bc2} \cite{pug2019}. 

Next, we will calculate the radiance of band $b$ in band 7 \cite{noaa_formula}:

\begin{equation}
    Rad_{b\_in\_b7} = \frac{fk_{1}}{\exp(\frac{fk_{2}}{BT \times bc_{2} +bc_{1}}) - 1}
\end{equation}

After that, we normalized every band into the domain of $[0, 1]$ to even cut down the size. Notice here we used {\tt dask.array} to store chunks instead of {\tt numpy}, which can cause bottleneck while calculating the data (\ref{lst:normalization}):

\begin{listing}[tb]%
\caption{Normalization}%
\label{lst:normalization}%
\begin{lstlisting}[language=Python]
def normalization(crop: da.Array) -> da.Array:
    stack_len, _, _ = crop.shape

    dif_max = da.nanmax(crop, axis=(1, 2))
    dif_min = da.nanmin(crop, axis=(1, 2))

    new_crop_stack = []
    for i in range(stack_len):
        curr = crop[i]
        new_crop_stack.append((curr - dif_min[i]) / (dif_max[i] - dif_min[i]))

    new_crop_stack = da.stack(new_crop_stack, axis=0)
    return new_crop_stack
\end{lstlisting}
\end{listing}

\begin{figure*}[t]\centering
    \centering
    \includegraphics[width=\textwidth]{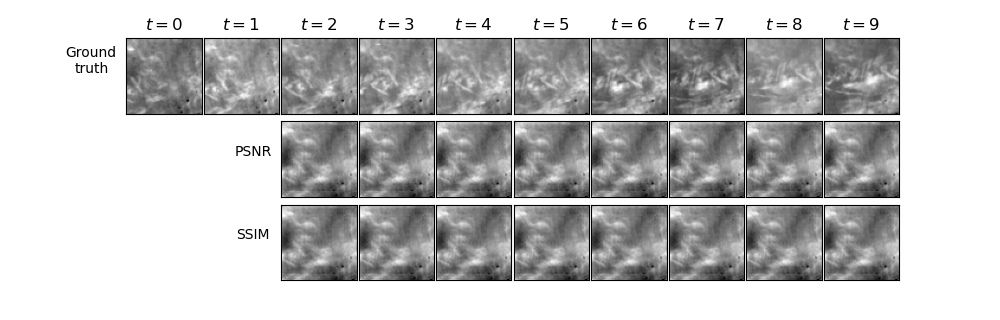}
    \caption{Ground truth (conditioning frames) and generated frames from our model.}
    \label{fig:generated}
\end{figure*}

We treated each band slice as a single channel for our image. Therefore a single "image" can have 10 "channels." To explore the data's stochastic characteristics, we can stack them to create a "video" with 12 frames in total {\tt video.shape = (12, 10, 256, 256)}. Finally, videos are compressed as {\tt .npz} files to reduce the size and increase the flexibility in storing.

\begin{figure} \centering
    \includegraphics[width=1.1\columnwidth]{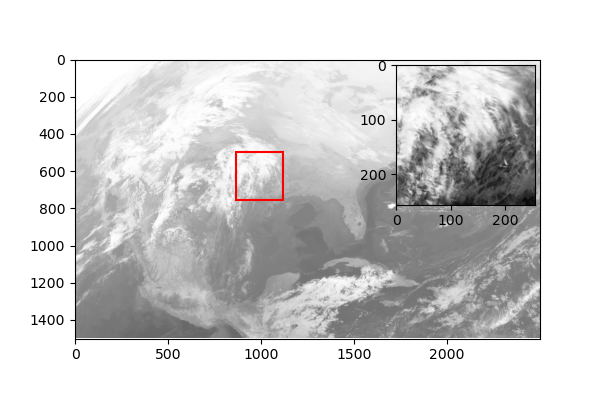}
    \caption{Band 7 and crop region}
    \label{fig:my_label}
\end{figure}

\begin{figure} \centering
    \centering
    \includegraphics[width=1.1\columnwidth]{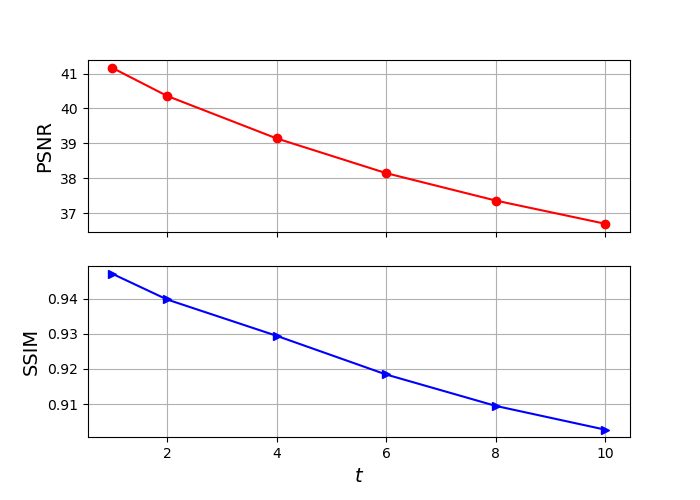}
    \caption{PSNR and SSIM scores with respect to $t$ the dataset.}
    \label{fig:graph}
\end{figure}

\begin{figure} \centering
    \centering
    \includegraphics[width=1.1\columnwidth]{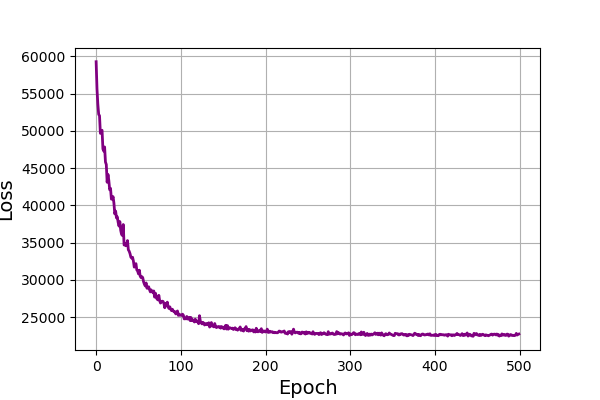}
    \caption{Loss vs. epochs over time}
    \label{fig:loss-e-epochs}
\end{figure}

\begin{figure} \centering
    \centering
    \includegraphics[width=1.1\columnwidth]{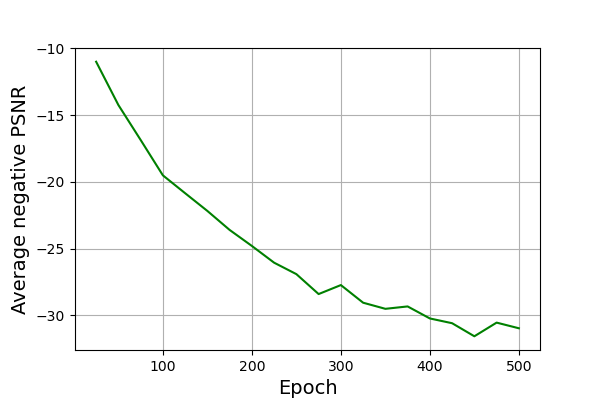}
    \caption{Average negative PSNR vs. epochs over time}
    \label{fig:avg-e-epochs}
\end{figure}

\section{Results and Discussion}
We trained the model on our GOES-16 dataset. The dataset is highly stochastic as the cloud can change its direction anytime. We set the size of the state-space variable $y$ and auxiliary variable $z$, both to $20$. For Euler step that shown in Equation (\ref{eq:2}), we set it to $2$. We used $2$ conditioning frames to generate the next $h$ frames, as shown in Figure (\ref{fig:graph}). Our method averages $40.43$ on PSNR and $0.934$ on SSIM. As discussed by \cite{villegas2017}, expanding network capacity can enhance the performance of such variational models. However, this is outside the scope of our work.

We challenge here the ODE inspiration of our model. Equation (\ref{eq:2}) amounts to learning a residual function $f_{z_{\lfloor t\rfloor + 1}}$ over $t \in [\lfloor t\rfloor, \lfloor t\rfloor + 1]$. We aim to test whether this dynamic is close to its continuous generalization:

\begin{equation}
\label{eq:11}
    \frac{dy}{dx} = f_{z_{\lfloor t\rfloor + 1}}(y)
\end{equation}

which is a piecewise ODE. To this end, we refine this Euler approximation during testing by using $\frac{\Delta t}{2}$; if this maintains the performance of our model, then the dynamic rule of the latter is close to the piecewise ODE, as shown in Figure (\ref{fig:graph}).

\section{Conclusion and Future Works}
We provide a unique dynamic latent model for stochastic video prediction that decouples frame synthesis and dynamics, unlike previous image-autoregressive models. This temporal model is based on residual updates of a tiny latent state and has outperformed RNN-based models. This confers numerous desired qualities on our strategy, including temporal economy and latent space interpretability. We empirically illustrate the proposed model's performance and benefits, which beats previous state-of-the-art approaches for stochastic video prediction. To the best of our knowledge, this is the first paper to offer a latent dynamic model that scales for video prediction. The suggested model is particularly innovative compared to current work on neural networks and ODEs for temporal modeling; it is the first residual model to scale to complex stochastic data such as videos.

We believe that the major principles of our method (state-space, residual dynamic, static content variable) may be applied to different models. We will supply a large amount of data for future studies, from GOES-16 (East) and GOES-17 (West), to provide the variety in wildfire circumstances. Furthermore, instead of employing ten bands, we may enhance our general characteristic of picture slices by mixing all visible and near-infrared bands with infrared to produce a full 16-band image. This gives the output a more distinct and realistic appearance, so the produced frames have more informative values.

\section{Acknowledgements}
We would like to thank all members, including students and staffs of 2022 Undergraduate Student Programmer at SSEC for helful discussions and comments, as well as William Roberts for his help to process the GOES-16 dataset.

\bibliography{main}
\end{document}